\documentclass[a4paper]{elsarticle}

\usepackage[numbers,sort&compress]{natbib}
\usepackage{amsmath}
\usepackage{array}
\usepackage{booktabs}
\usepackage{arydshln}

\def\tsc#1{\csdef{#1}{\textsc{\lowercase{#1}}\xspace}}
\tsc{WGM}
\tsc{QE}

\begin{document}
\begin{sloppypar}

\let\WriteBookmarks\relax
\def\floatpagepagefraction{1}
\def\textpagefraction{.001}

\title[mode=title]{Contrastive Variational Information Bottleneck for Aspect-Based Sentiment Analysis} 

\shorttitle{Contrastive Variational Information Bottleneck for Aspect-Based Sentiment Analysis}    

\shortauthors{M. Chang et al.}



%

\author[1,2]{Mingshan Chang}[style=Chinese]
\ead{ms.chang@siat.ac.cn}
\credit{Conceptualization, Methodology, Software, Writing – original draft}

\author[1]{Min Yang}[style=Chinese]
\cormark[1]
\ead{min.yang@siat.ac.cn}
\credit{Conceptualization, Methodology, Supervision, Writing – review \& editing}

\author[1]{Qingshan Jiang}[style=Chinese]
\ead{qs.Jiang@siat.ac.cn}
\credit{Writing – review \& editing}

\author[3]{Ruifeng Xu}[style=Chinese]
\ead{xuruifeng@hit.edu.cn}
\credit{Writing – review \& editing}

\affiliation[1]{
organization={Shenzhen Institutes of Advanced Technology},
addressline={Chinese Academy of Sciences}, 
city={Shenzhen},
country={China}
}

\affiliation[2]{
organization={University of Chinese Academy of Sciences},
city={Beijing},
country={China}
}

\affiliation[3]{
organization={School of Computer Science and Technology},
addressline={Harbin Institute of Technology (Shenzhen)}, 
city={Shenzhen},
country={China}
}

\cortext[1]{Corresponding author}



\begin{abstract}
Deep learning techniques have dominated the literature on aspect-based sentiment analysis (ABSA), achieving state-of-the-art performance. However, deep models generally suffer from spurious correlations between input features and output labels, which significantly hurts the robustness and generalization capability. 
In this paper, we propose to reduce spurious correlations for ABSA, via a novel \underline{C}ontrastive \underline{V}ariational \underline{I}nformation \underline{B}ottleneck framework (called CVIB).
The proposed CVIB framework is composed of an original network and a self-pruned network, and these two networks are optimized simultaneously via contrastive learning. Concretely, we employ the Variational Information Bottleneck (VIB) principle to learn an informative and compressed network (self-pruned network) from the original network, which discards the superfluous patterns or spurious correlations between input features and prediction labels. Then, self-pruning contrastive learning is devised to pull together semantically similar positive pairs and push away dissimilar pairs, where the representations of the anchor learned by the original and self-pruned networks respectively are regarded as a positive pair while the representations of two different sentences within a mini-batch are treated as a negative pair.
To verify the effectiveness of our CVIB method, we conduct extensive experiments on five benchmark ABSA datasets. The experimental results show that our approach achieves better performance than the strong competitors in terms of overall prediction performance, robustness, and generalization.
\end{abstract}




\maketitle
\section{Introduction}
\label{introduction}
With the growing abundance of opinion-rich content on the Web, aspect-based sentiment analysis (ABSA), which aims to identify the sentiment polarity of a sentence towards a given aspect, has attracted great attention from both academic and industrial communities.
Conventional ABSA methods mainly employ supervised machine learning techniques involving various hand-crafted features such as syntactic features \cite{negi2014insight}, parse trees \cite{pekar2014ubham}, and lexical features \cite{negi2014insight}, to predict the sentiment polarity. However, the process of feature engineering is labor-intensive and the hand-crafted features cannot be adapted to new domains easily.

In recent years, deep learning techniques have emerged as the mainstream in the literature on ABSA. 
Prominent deep neural networks can be trained end-to-end to automatically learn semantically distinguishable representations for both the aspect and context without manual annotation.
To capture crucial sentiment information related to the target aspect, various attention mechanisms \cite{wang2016attention, tang-etal-2016-aspect, Yang_Tu_Wang_Xu_Chen_2017, ma2017interactive, he-etal-2018-effective, fan2018multi, li-etal-2018-hierarchical} have been proposed to model the interactions between the aspect and its context. 
Subsequently, several studies leveraged syntactic knowledge and graph neural networks to capture syntax-aware features for the target aspect explicitly \cite{huang-carley-2019-syntax, zhang-etal-2019-aspect, sun2019aspect, wang2020relational, tian-etal-2021-aspect, WU2022107736, LIANG2022107643}, to improve the performance of the ABSA models. 
More recently, there has been a notable application of pre-trained language models (PLMs) such as BERT \cite{devlin2018bert} and RoBERTa \cite{liu2019roberta} to learn effective task-specific representations \cite{song2019attentional, jiang-etal-2019-challenge, wang2020relational, dai-etal-2021-syntax, zhang-etal-2022-incorporating, YOU2022109511}, yielding state-of-the-art results for ABSA.

\begin{figure}[tp]
\centering
\includegraphics[width=81mm]{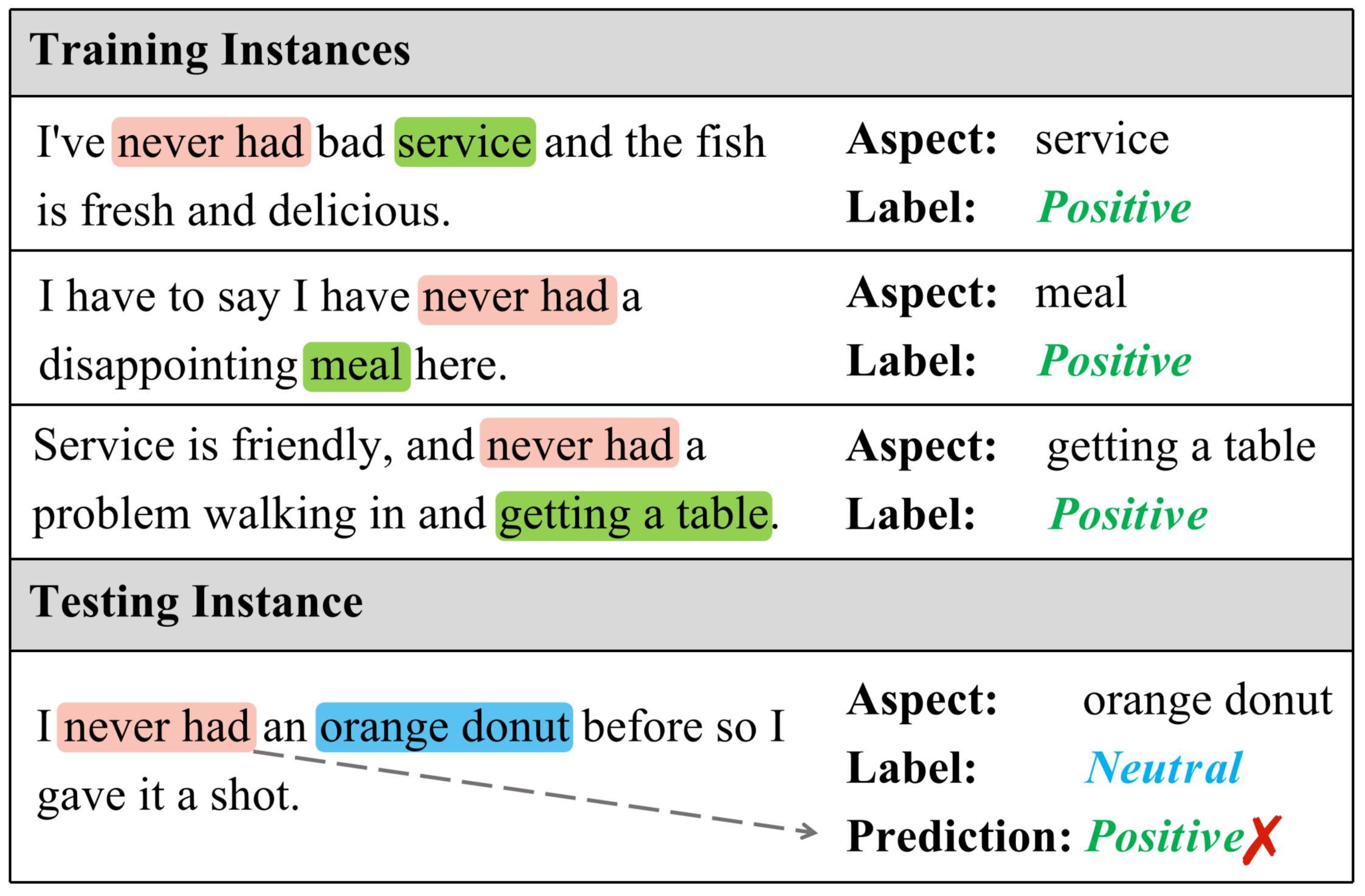}
\caption{Examples of the spurious correlation between the context words ``\textit{never had}'' and the sentiment label ``\textsc{Positive}'' in the training corpus, and the spurious correlation does not hold for the testing instance.}
\label{fig:example}
\end{figure}

Despite the remarkable progress, deep ABSA models are notoriously brittle to learn statistically spurious correlations between learned patterns and prediction labels \cite{absa_survey_zhang_2022,xing-etal-2020-tasty}. The spurious correlations are defined as the superficial feature patterns that hold for most training examples but are not inherent to the task of interest.
As shown in Fig. \ref{fig:example}, we provide an example of the online restaurant review to illustrate the spurious correlation problem that existed in ABSA. Based on our empirical observation, in the training phase, the deep models tend to learn the high correlations between the context words ``\textit{never had}'' and the sentiment polarity label ``\textsc{Positive}'' without taking the aspect words into consideration. That is, the models may be ``right for the wrong reasons'' due to the reliance
on the spurious correlation between the presence of context words ``\textit{never had}'' and the label ``\textsc{Positive}''.
Consequently, the learned sentiment classifier would fail to predict the correct sentiment label ``\textsc{Neutral}'' for the testing instance where the spurious correlation does not hold.
Under such an inductive bias, a deep ABSA model usually 
learn sub-optimal feature representations, especially for the under-represented classes (the long-tail samples), which significantly hurts the robustness and generalization capability.  

One possible solution to mitigate the spurious correlation problem in ABSA is to prune spurious features from an information bottleneck perspective. The key idea is to automatically learn the essential contextual representations that contain minimal relevant information about the inputs while preserving sufficient information for label prediction, making the deep models more robust and generalized against statistically spurious correlations. In particular, the 
Variational Information Bottleneck (VIB) is a technique in information theory \cite{tishby2000information} for suppressing irrelevant features, which minimizes the mutual information (MI) between the inputs and internal representations while maximizing the MI between the outputs and the representations. 
We hypothesize that VIB can mitigate the overfitting problem and provide an advantageous inductive bias for the target tasks, thus resulting in better robustness and generalization to challenging out-of-domain data. 
However, VIB is computationally intractable due to the non-differentiable categorical sampling, severely limiting the application of the VIB principle in ABSA. 
In addition, deep ABSA models generally struggle to characterize challenging long-tail samples. Devising a strategy to extract the inherent characteristics of each class and distinguish different classes is a potentially fruitful research direction for improving the robustness and generalization capability.

To tackle the aforementioned challenges, we propose a \underline{C}ontrastive \underline{V}ariational \underline{I}nformation \underline{B}ottleneck framework (called CVIB) to mitigate the spurious correlation problem for the ABSA task, aiming at improving the robustness and generalization capability of the deep ABSA method. The proposed CVIB framework prevents the deep ABSA model from capturing spurious correlation even without prior knowledge of the biased information by simultaneously considering information compression and retention from the information-theoretic perspective. Concretely, CVIB is composed of an original network and a self-pruned network, which are optimized simultaneously via contrastive learning. The self-pruned network is learned adaptively from the original network based on the \underline{V}ariational \underline{I}nformation \underline{B}ottleneck principle, which is expected to discard the spurious correlations while preserving sufficient information about the sentiment labels.
A self-pruning contrastive loss is then devised to optimize the two networks and improve the separability of all the classes, which narrows the distance between the representations of each anchor produced by the self-pruned and original networks while pushing apart the distance between the representations of different instances within a mini-batch. Consequently, the self-pruned network reduces the spurious correlations, making it easier for the ABSA classifier to avoid overfitting. The main contributions of this paper are listed as follows:
\begin{itemize}
\item We propose a CVIB framework to reduce spurious correlations between input features and output labels without prior knowledge of such correlations, which improves the robustness and generalization capability of the deep ABSA model by taking advantage of both VIB and contrastive learning. 
\item We devise self-pruning contrastive learning to extract truly essential semantically relevant features and effectively generalize to long-tail instances by learning the inherent class characteristics.
\item We conduct extensive experiments on five benchmark datasets, showing that the proposed CVIB method achieves better performance than the strong baselines in terms of overall prediction performance, robustness, and generalization.
\end{itemize}

\begin{figure*}[tp]
\centering
\includegraphics[width=177mm]{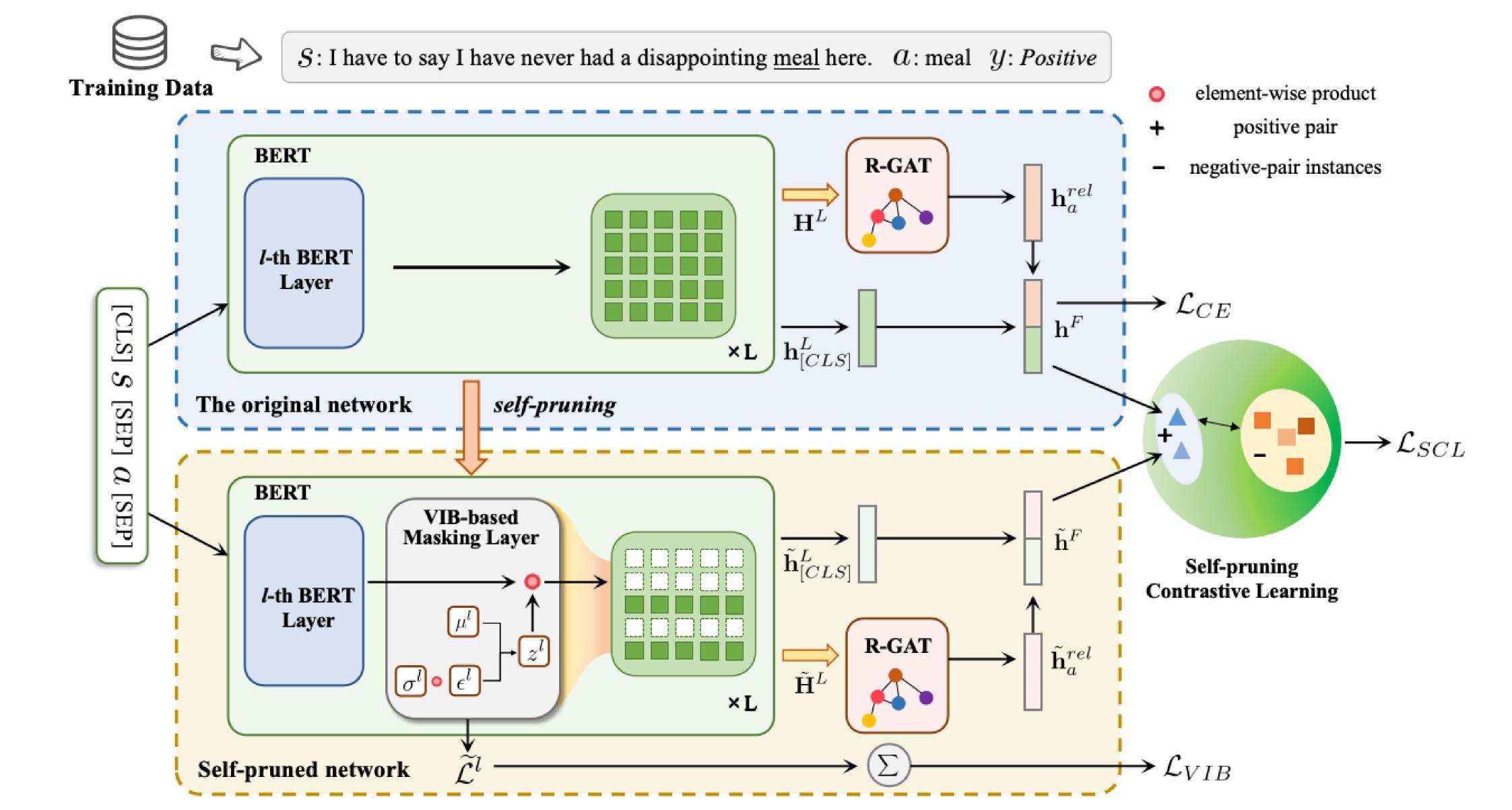}
\caption{\label{fig:CVIB} The architecture of the proposed CVIB framework for ABSA. CVIB is composed of an original network and a self-pruned network, where the self-pruned network is learned adaptively from the original network based on the Variational Information Bottleneck (VIB) principle. The self-pruned network is expected to discard the spurious correlations between input features and output prediction, which is used for inference.}
\end{figure*}

\section{Related Work}\label{related works}
\subsection{Aspect-based Sentiment Analysis Using Deep Learning}
Aspect-based sentiment analysis (ABSA) stands as a vital task in the field of sentiment analysis. The goal of ABSA is to automatically detect the sentiment polarity of the input sentence towards a given aspect. Currently, deep neural networks have become the predominant methods in ABSA due to their impressive performance.
Early deep ABSA methods concentrated on crafting diverse attention mechanisms to implicitly capture the semantic relationship between the given aspect and its context by learning attention-based representations \cite{wang2016attention, ma2017interactive, Yang_Tu_Wang_Xu_Chen_2017, he-etal-2018-effective, li-etal-2018-hierarchical, fan2018multi, Wang_Lu_2018}.
\citet{wang2016attention} introduced attention-based LSTMs to capture relevant sentiment information from the context concerning the target aspect. 
 \citet{Wang_Lu_2018} proposed a segmentation attention mechanism to capture the structural dependencies between the target aspect and its context words.
 \citet{ma2017interactive} devised an interactive attention mechanism to interactively learn the attention-aware representations of the target aspect and its context. 

Another research trend involves explicitly capturing syntax-aware features for the target aspect by leveraging syntactic knowledge and graph neural networks \cite{huang-carley-2019-syntax, zhang-etal-2019-aspect, sun2019aspect, wang2020relational, tian-etal-2021-aspect, li-etal-2021-dual-graph, LIANG2022107643, WU2022107736, LU2022109840}. The fundamental concept behind these methods is to construct the syntax dependency tree for a sentence and convert it into a graph. Subsequently, graph convolutional networks (GCNs) \cite{zhang-etal-2019-aspect, sun2019aspect, tian-etal-2021-aspect, li-etal-2021-dual-graph, LIANG2022107643, LU2022109840} or graph attention networks (GATs) \cite{huang-carley-2019-syntax, wang2020relational, WU2022107736} are employed to aggregate sentiment information from the neighboring context nodes to the target aspect node. 

More Recently, the pre-trained language models (PLMs), such as BERT \cite{devlin2018bert} and RoBERTa \cite{liu2019roberta}, have been applied to ABSA, achieving state-of-the-art performance \cite{song2019attentional, jiang-etal-2019-challenge, wang2020relational,Wu_Ong_2021, zhang-etal-2022-incorporating, dai-etal-2021-syntax, YOU2022109511}. These methods either incorporated BERT/RoBERTa as an embedding layer \cite{jiang-etal-2019-challenge, wang2020relational} or fine-tuned specific BERT/RoBERTa-based models with a classification layer \cite{xu-etal-2019-bert, Wu_Ong_2021, zhang-etal-2022-incorporating}. In this way, extensive linguistic knowledge learned from large textual corpora can be exploited to improve the performance of ABSA.
However, these deep ABSA models are susceptible to spurious correlations between input features and output labels, resulting in poor robustness and generalization capability. In this paper, our main emphasis is on reducing spurious correlations to learn a more robust and generalizable ABSA model.

\subsection{Spurious Correlation Reduction in NLP}
Deep neural networks, although powerful, generally present a tendency to learn spurious correlations and suffer from the overfitting issue on manually-annotated datasets \cite{jia-liang-2017-adversarial, gururangan-etal-2018-annotation, kaushik-lipton-2018-much, sanchez-etal-2018-behavior, mccoy-etal-2019-right, niven-kao-2019-probing}. This challenge is evident across a wide range of NLP tasks, including natural language inference \cite{gururangan-etal-2018-annotation, mccoy-etal-2019-right}, question answering \cite{jia-liang-2017-adversarial} and reading comprehension \cite{kaushik-lipton-2018-much}, making the trained models unstable and cannot generalize well to out-of-distribution \cite{Ming_Yin_Li_2022, NEURIPS2022_f0e91b13} or open-set data \cite{pmlr-v139-fang21c} in real-world scenarios.

To address these challenges, some studies proposed to explicitly reduce the spurious correlations present in the original datasets by removing the data biases \cite{zellers-etal-2019-hellaswag, kaushik2019learning, sakaguchi2020winogrande, nie-etal-2020-adversarial, Wang_Culotta_2021, wu-etal-2022-generating}. For instance, \citet{zellers-etal-2019-hellaswag} employed adversarial filtering methods to create debiased datasets, mitigating spurious artifacts present in the original training data. \citet{nie-etal-2020-adversarial} generated additional samples for the original training samples that exhibited vulnerability to spurious correlations through an iterative human-in-the-loop process. 
In parallel, several model-centered approaches \cite{clark-etal-2020-learning, karimi-mahabadi-etal-2020-end, utama-etal-2020-mind, sanh2021learning, du-etal-2021-towards, Du_Tang_Fu_Hu_2022} built models dedicated to capturing spurious features in the training data and utilized re-weighting strategies to train debiased models based on the detected spurious correlations.
For example, \citet{clark-etal-2020-learning} introduced a low-capacity model to capture shallow patterns and down-weighted them in the training objective via ensemble learning, facilitating the learning of a more robust model. \citet{stacey-etal-2020-avoiding} devised a bias-only classifier to capture spurious features and deterred the hypothesis encoder from learning them, which in turn updated the classifier through an adversarial learning approach.

More recently, there has been an increasing focus on mitigating overfitting issues from an information-theoretic perspective \cite{tian-etal-2021-embedding, lovering2021predicting, pmlr-v139-zhou21g, mahabadi2021variational}. For example, \citet{mahabadi2021variational} employed the variational information bottleneck (VIB) principle to eliminate irrelevant features from learned representations, enhancing generalization to out-of-domain data. \citet{tian-etal-2021-embedding} introduced disentangled semantic representations for minority samples (i.e., \textit{long-tail} samples), utilizing mutual information to alleviate surface patterns prevalent in majority samples. 

In this paper, we employ the variational information bottleneck (VIB) principle to automatically reduce spurious or redundant information. Additionally, we devise a self-pruning contrastive learning to extract more semantically relevant features, improving the separability of all sentiment classes. Our goal is to reduce spurious correlations to improve the robustness and generalization capability of ABSA. Furthermore, we offer a potential solution for out-of-distribution detection and open-set learning from an information-theoretical perspective.

\section{Methodology}\label{methodology}
We assume there are $N$ instances in the training set, where each instance $x$ contains a context text $s=\{w^s_i\}_{i=1}^{n}$ with $n$ words and a target aspect $a=\{w^a_i\}_{i=1}^{m}$ with $m$ words. $w^s_i$ (or $w^a_i$) denotes the $i$-th word in the context (or target aspect). Each instance $x$ has a sentiment category label $y\in \{1, \ldots, C\}$, where $C$ stands for the number of sentiment categories. The goal of ABSA is to predict the sentiment polarity $\hat{y}$ towards the target aspect given the input instance $x$.

Fig. \ref{fig:CVIB} illustrates the overview of the proposed CVIB framework. CVIB is composed of an original network $mathcal{M}_{\theta_1}$ and a self-pruned network $\mathcal{M}_{\theta_2}^{(p)}$, which are optimized simultaneously via contrastive learning. Here, $\theta_1$ and $\theta_2$ represent the parameter sets of the original and self-pruned networks, respectively. In particular, the self-pruned network is compressed adaptively from the original network based on the Variational Information Bottleneck principle, which is expected to discard the spurious correlations between input features and output prediction. A self-pruning contrastive loss is then devised to optimize the two networks and improve the separability of all the sentiment classes. Next, we will describe the original network, the self-pruned network, and the self-pruning contrastive learning in detail.

\subsection{The Original Network}
Inspired by the remarkable success of pre-trained language models (PLMs),  we employ BERT \cite{devlin2018bert} as our base text encoder to learn the semantic representations of the target aspect and its context for sentiment prediction. In addition, following the previous work \cite{wang2020relational}, we also leverage a relational graph attention network to capture the aspect-oriented syntactic structure of the sentence. 

\subsubsection{Base BERT Encoder} 
Given a sequence of context words $s=\{w^s_i\}_{i=1}^n$ and the corresponding target $a=\{w^a_i\}_{i=1}^{m}$,  we adopt the pre-trained language model BERT \cite{devlin2018bert} as the base text encoder and take the formatted sequence ${\mathbf{x}=(\textsc{[CLS]} s \textsc{[SEP]} a \textsc{[SEP]})}$ as the input, where the special tokens \textsc{[CLS]} and \textsc{[SEP]} represent the classification token and the separation token respectively. The task-relevant features can be captured through successive layers of BERT, in which the ${l}$-th layer can be calculated as:
\begin{flalign}
    & \mathbf{H}^{1}= \mathrm{BERTLayer}\left(\mathbf{x}\right)\\
    & \mathbf{H}^{l}=\mathrm{BERTLayer}\left(\mathbf{H}^{l-1}\right) &
\end{flalign}
where $\mathbf{H}^{l}=[\mathbf{h}_{1}^{l},\dots,\mathbf{h}_{m+n+3}^{l}]$ represents the output hidden representations of the ${l}$-th BERT layer ($l\in \{2,\dots,L\}$) and $L$ is the number of the BERT layers. $\mathbf{h}_{i}^{l}$ denotes the $i$-th item in $\mathbf{H}^{l}$. 

\subsubsection{Relational Graph Attention Network} 
We employ a relational graph attention network (denoted as R-GAT) \cite{wang2020relational} to capture the aspect-oriented syntactic dependency, which first converts different dependency relations into embeddings through an embedding layer and then incorporates them in computing multi-head relational attention to obtaining syntax-aware representations $\mathbf{H}^{\textit{rel}}$. For simplicity, we denote the computation of R-GAT as:  
\begin{flalign}
   & \mathbf{H}^{{rel}}=\mathrm{RGAT}\left(\mathbf{H}^{L}\right) &
\end{flalign}
where ${\mathbf{H}^{L}}$ and ${\mathbf{H}^{rel}}$ denote the output representations from the BERT encoder and the R-GAT network, respectively.

\subsubsection{Aspect-based Sentiment Prediction} 
The final representation $\mathbf{h}^{F}$ is formed by concatenating the output representation ${\mathbf{h}^{L}_{[CLS]}}$ of the first token \textsc{[CLS]} from the BERT encoder and the output syntactic-aware representation ${\mathbf{h}^{rel}_{a}}$ of the target aspect $a$ from the R-GAT network, which is denoted as:
\begin{flalign}
    & \mathbf{h}^{F}={\mathbf{h}_a^{rel} \Vert \, \mathbf{h}_{[CLS]}^L} &
\end{flalign}
Then, we feed ${\mathbf{h}^{F}}$ into a multi-layer perceptron (MLP) layer followed by a softmax layer to predict the sentiment distribution:
\begin{flalign}   & \hat{\mathbf{y}}=\mathrm{softmax}\left(W^o{\mathrm{MLP}\left(\mathbf{h}^{F}\right)}+b^o\right) &
\end{flalign}
where ${\hat{\mathbf{y}}}$ is the predicted sentiment distribution. $W^o$ and $b^o$ are learnable parameters.
Given a corpus with $N$ training samples $(x_i, y_i)_{i=1}^N$, the parameters of the sentiment classifier are trained to minimize the standard cross-entropy loss function:
\begin{flalign}
   & \mathcal{L}_{CE}\left(\theta_1\right)=-\sum_{i=1}^{N}\mathbf{y}_{i}\mathrm{log}\,\left(\hat{\mathbf{y}}_{i}\right) &
\end{flalign}
where ${{\mathbf{y}}_{i}^{j}}$ and ${\hat{\mathbf{y}}_{i}^{j}}$ are the ground-truth and predicted sentiment probabilities, respectively.  ${\theta}_1$ represents the set of learnable parameters of the original network $\mathcal{M}_{\theta_{1}}$.

\subsection{The Self-pruned Network based on VIB}
\label{subsec: Self-pruned Network}
Spurious correlations are very common in deep models, especially when the ABSA classifier is over-parameterized. Spurious correlations could hurt the stability and generality of the ABSA classifier when deployed in practice. In this paper, we learn a self-pruned network adaptively from the original network based on the Variational Information Bottleneck (VIB) principle, which is expected to reduce the spurious correlations while preserving sufficient information for output prediction.
To be specific, we aim to remove irrelevant or redundant information from hidden representations layer by layer. We learn a compressed self-pruned network $\mathcal{M}_{\theta_2}^{(p)}$ from the original network $\mathcal{M}_{\theta_1}$ via the self-pruning technique, improving the robustness and generalization capability.

\subsubsection{Variational Information Bottleneck}
The goal of VIB is to learn a compressed representation $\tilde{\mathbf{H}}^{l}$ while retaining sufficient information in $\mathbf{H}^{l}$ required for prediction.
For the ${l}$-th BERT layer, we minimize the mutual information $I\left(\tilde{\mathbf{H}}^{l-1},\tilde{\mathbf{H}}^{l}\right)$ between the input hidden states ${\tilde{\mathbf{H}}^{l-1}}$ and the output hidden states ${\tilde{\mathbf{H}}^{l}}$ to maximally reduce irrelevant information across BERT layers. Meanwhile, IB is also expected to maximize the mutual information $I\left(\tilde{\mathbf{H}},\mathbf{y}\right)$ between the compressed output hidden states ${\tilde{\mathbf{H}}^{l}}$ and the target label $\mathbf{y}$ so as to preserve sufficient task-relevant information for accurate prediction of $\mathbf{y}$. Mathematically, the layer-wise training objective of the self-pruned network $\mathcal{M}_{\theta_2}^{(p)}$ is to minimize the following loss:
\begin{flalign}
    & \mathcal{L}^{l}\left(\theta_2\right)=\beta^{l} I\left(\tilde{\mathbf{H}}^{l-1},\tilde{\mathbf{H}}^{l}\right)-I\left(\tilde{\mathbf{H}}^{l},\mathbf{y}\right) &
\label{eq:IB}   
\end{flalign}

where ${I(\cdot)}$ indicates mutual information between two  variables. The hyper-parameter ${\beta^{l}}$ controls compression-accuracy trade-off.

Unfortunately, mutual information is computationally intractable for general deep neural network architectures, making the optimization of Eq. (\ref{eq:IB}) difficult. To conquer this challenge, the Variational Information Bottleneck (VIB) principle \cite{alemi2016deep,fabius2015variational} invokes tractable variational bounds as efficient approximations for the optimization objective.
Similarly, we derive a variational upper bound for ${\mathcal{L}^{l}}$:
\begin{flalign}
\label{eq:VIB bound}
\tilde{\mathcal{L}}^{l}\left(\theta_2\right) & = \beta^{l} \mathbb{E}_{ \mathbf{x},\mathbf{y},\tilde{\mathbf{H}}^{1:l-1}}\left[\mathbb{KL}\left[p\left(\tilde{\mathbf{H}}^{l}|\tilde{\mathbf{H}}^{l-1}\right)\Vert q\left(\tilde{\mathbf{H}}^{l}\right)\right]\right] \nonumber \\
& - \mathbb{E}_{ \mathbf{x},\mathbf{y},\tilde{\mathbf{H}}^{L}}\left[\mathrm{log}\, q\left(\mathbf{y} | \tilde{\mathbf{H}}^{L}\right)\right] &
\end{flalign}
where ${\tilde{\mathbf{H}}^{1:l-1} \triangleq \{\tilde{\mathbf{H}}^j\}_{j=1}^{l-1}}$. $q\left(\tilde{\mathbf{H}}^{l}\right)$ and ${q\left(\mathbf{y} | \tilde{\mathbf{H}}^{L}\right)}$ are parametric variational approximation to $p\left(\tilde{\mathbf{H}}^{l}\right)$ and ${p\left(\mathbf{y} | \tilde{\mathbf{H}}^L\right)}$, respectively. Inside the expectation, the first KL divergence term aims to reduce superfluous information between adjacent layers, while the second term focuses on retaining sufficient task-relevant information for accurate sentiment prediction.

\subsubsection{VIB-based Masking Layer} To reduce superfluous information from hidden representations, we add a masking layer into each adjacent BERT layer. Concretely, we apply a  mask (denoted as $\mathbf{z}^l$) to the output representations of each BERT layer $l$. The mask $\mathbf{z}^l$ is shared by the hidden vectors within each BERT layer, while different BERT layers have different masks. Formally, for the $l$-th BERT layer, we calculate the mask $\mathbf{z}^l$ and the compressed hidden representation $\tilde{\mathbf{H}}^l$ as follows: 
\begin{flalign}
    & \tilde{\mathbf{H}}^{l}=\mathbf{z}^l \odot f^l(\tilde{\mathbf{H}}^{l-1}),\quad  \mathbf{z}^l={\mu}^l+{{\epsilon}}^{l}\odot {\sigma}^l &
\end{flalign}
\noindent where $\odot$ denotes element-wise multiplication. $\mu^l$ and $\sigma^l$ are learnable vectors. ${\epsilon^{l}}$ is a shared vector sampled from the normal distribution $\mathcal{N}\left(0,I\right)$.  ${f^l\left(\cdot\right)}$ indicates the ${l}$-th BERT layer. With these definitions, the conditional layer-wise distribution ${p\left(\tilde{\mathbf{H}}^{l}|\tilde{\mathbf{H}}^{l-1}\right)}$ can be specified as:
\begin{small}
\begin{flalign}
& p\left(\tilde{\mathbf{H}}^{l}|\tilde{\mathbf{H}}^{l-1}\right)= \mathcal{N}\left(\tilde{\mathbf{H}}^{l};f^l(\tilde{\mathbf{H}}^{l-1})\odot {\mu}^l, \mathrm{diag}[f^l(\tilde{\mathbf{H}}^{l-1})^2\odot \left({{\sigma}^l}\right)^2]\right) &
\label{eq:layer-wise distribution}
\end{flalign}
\end{small}

In addition, we assume that ${q\left(\tilde{\mathbf{H}}^{l}\right)}$ also follows a Gaussian distribution  ${\mathcal{N}\left(0,\mathrm{diag}[{\xi}^l]\right)}$ with a zero mean value and a variance vector ${\xi}^{l}$, which can be learned out of the model.
Then, we take the defined $p\left(\tilde{\mathbf{H}}^{l}|\tilde{\mathbf{H}}^{l-1}\right)$ and ${q\left(\tilde{\mathbf{H}}^{l}\right)}$ into the KL term of Eq. (\ref{eq:VIB bound}) and calculate the KL divergence between two Gaussian distributions with the standard formula:
\begin{small}
\begin{flalign}
\label{eq:kl_1} 
 &\mathbb{E}_{\tilde{\mathbf{H}}^{l-1}}\left[\mathbb{KL}\left[p\left(\tilde{\mathbf{H}}^{l}|\tilde{\mathbf{H}}^{l-1}\right)\Vert q\left(\tilde{\mathbf{H}}^{l}\right)\right] \right] = \nonumber\\
 &\frac{1}{2}\mathbb{E}_{\tilde{\mathbf{H}}^{l-1}}\sum_{j}^{|\tilde{\mathbf{H}}|}[\frac{\left((\mu_{j}^{l})^2+(\sigma_{j}^{l})^2\right)\cdot f_j^l(\tilde{\mathbf{H}}^{l-1})^2}{\xi^l_j} - \mathrm{log}\frac{(\sigma_{j}^{l})^2\cdot f_j^l(\tilde{\mathbf{H}}^{l-1})^2}{\xi^l_j}-1] &
\end{flalign}
\end{small}
where ${|\tilde{\mathbf{H}}|}$ is the dimension of hidden vectors $\tilde{\mathbf{H}}^{l}$. ${\mu_{j}^{l}}$, ${\sigma_{j}^{l}}$ are the ${j}$-th element of the corresponding vectors. 

Then, we calculate the gradient of Eq. (\ref{eq:kl_1}) and set it equal to zero to find the optimal value of $\xi_j^l$, which is denoted as $\xi_j^{l \ast}$:
\begin{flalign}
& \mathbb{E}_{\tilde{\mathbf{H}}^{l-1}}[-\frac{\left((\mu_{j}^{l})^2+(\sigma_{j}^{l})^2\right)\cdot f_j^l(\tilde{\mathbf{H}}^{l-1})^2}{(\xi_j^{l \ast})^2}+\frac{1}{\xi_j^{l \ast}}]  = 0  \nonumber &  \\
& \xi_j^{l \ast}=\left((\mu_{j}^{l})^2+(\sigma_{j}^{l})^2\right)\cdot \mathbb{E}_{\tilde{\mathbf{H}}^{l-1}}[f_j^l(\tilde{\mathbf{H}}^{l-1})^2] \nonumber &  
\end{flalign}

Based on the above formula, we can observe that if an arbitrary element of ${{\xi}^l}$ is learned to be close to zero, the corresponding element of $p(\tilde{\mathbf{H}}^{l}|\tilde{\mathbf{H}}^{l-1})$ will be pushed towards a degenerate Dirac-delta and can be further pruned. Here, degenerate Dirac-delta means a deterministic distribution and takes only a single value (i.e., zero). 

For simplicity, we denote $g_j^l=\mathbb{E}_{\tilde{\mathbf{H}}^{l-1}}[f_j^l(\tilde{\mathbf{H}}^{l-1})^2]$ and take the expression of $\xi_j^{l \ast}$ back into Eq. (\ref{eq:kl_1}). In this way, we can obtain:\
\begin{flalign}
\label{eq:kl_2}
 & \mathrm{inf}_{{\xi^l}\succ 0}\,\mathbb{E}_{\tilde{\mathbf{H}}^{l-1}}
 \left[\mathbb{KL}\left[p\left(\tilde{\mathbf{H}}^{l}|\tilde{\mathbf{H}}^{l-1}\right)\Vert q\left(\tilde{\mathbf{H}}^{l}\right)\right] \right] \nonumber\\
& = \frac{1}{2}\mathbb{E}_{\tilde{\mathbf{H}}^{l-1}}\sum_{j}^{|\tilde{\mathbf{H}}|}[\mathrm{log}\frac{(\mu_{j}^{l})^2+(\sigma_{j}^{l})^2}{(\sigma_{j}^{l})^2}+\mathrm{log}\frac{g_j^l}{f_j^l(\tilde{\mathbf{H}}^{l-1})^2}] \nonumber\\
&= \frac{1}{2}\sum_{j}^{|\tilde{\mathbf{H}}|}[\mathrm{log}\left(1+\frac{(\mu_{j}^{l})^2}{(\sigma_{j}^{l})^2}\right)+\mathrm{log}g_j^l-\mathbb{E}_{\tilde{\mathbf{H}}^{l-1}}[\mathrm{log}f_j^l(\tilde{\mathbf{H}}^{l-1})^2]] \nonumber \\
&= \frac{1}{2} \sum_{j}^{|\tilde{\mathbf{H}}|}[\mathrm{log}\left(1+\frac{(\mu_{j}^{l})^2}{(\sigma_{j}^{l})^2}\right)+\psi^l_j] &
\end{flalign}
where 
\begin{flalign}
& {\psi}^{l}_{j}\triangleq \mathrm{log}\,\mathbb{E}_{\tilde{\mathbf{H}}^{l-1}}[f_j^l(\tilde{\mathbf{H}}^{l-1})^2] -\mathbb{E}_{\tilde{\mathbf{H}}^{l-1}}[\mathrm{log}f_j^l(\tilde{\mathbf{H}}^{l-1})^2] &
\end{flalign}
and according to Jensen's inequality, the value of ${\psi}^{l}_{j}$ is positive and close to zero when the variance of ${p\left(\tilde{\mathbf{H}}^{l-1}\right)}$ is small. Thus, it can be removed without affecting the results.

Then, the KL term in Eq. (\ref{eq:VIB bound}) has a tractable and closed-form approximation, which further simplifies  ${\tilde{\mathcal{L}}^{l}}$ defined in Eq. (\ref{eq:VIB bound}) as follows:
\begin{small}
\begin{flalign}
& \tilde{\mathcal{L}}^{l}\left(\theta_2\right) =\beta^{l}\sum_{j=1}^{{|\tilde{\mathbf{H}}|}}\mathrm{log}\left(1+\frac{(\mu_{j}^{l})^2}{(\sigma_{j}^{l})^2}\right) - \mathbb{E}_{\mathbf{x},\mathbf{y},\tilde{\mathbf{H}}^L}\left[\mathrm{log}\,q\left(\mathbf{y}|\tilde{\mathbf{H}}^L\right)\right] &
\label{eq:layer-wise VIB loss}
\end{flalign}
\end{small}

Therefore, the objective function of the self-pruned network is computed by summing up the loss $\tilde{\mathcal{L}}^{l}\left(\theta_2\right)$ of all BERT layers:
\begin{small}
\begin{flalign}
& {\mathbf{\mathcal{L}}}_{VIB}\left(\theta_2\right) =\sum_{l=1}^L\beta^{l}\sum_{j=1}^{|\tilde{\mathbf{H}}|}\mathrm{log}\left(1+\frac{{(\mu_{j}^{l})}^2}{{(\sigma_{j}^{l})}^2}\right) -L\mathbb{E}_{\mathbf{x},\mathbf{y},\mathbf{H}^L}\left[\mathrm{log}\,q\left(\mathbf{y}|\tilde{\mathbf{H}}^L\right)\right] &
\label{eq:final VIB loss}
\end{flalign} 
\end{small}
where ${\theta_2}$ represents the set of learnable parameters of the self-pruned network ${\mathcal{M}_{\theta_2}^{(p)}}$.

Consequently, the masking vectors $\{\mathbf{z}^l\}_{l=1}^{L}$ can be learned to reduce spurious information from hidden representations during the training process. To be specific, while minimizing the inter-layer mutual information $I\left(\tilde{\mathbf{H}}^{l-1},\tilde{\mathbf{H}}^{l}\right)$, we use the ratio $\boldsymbol{\alpha}_{j}^{l}=(\mu_{j}^{l})^2/(\sigma_{j}^{l})^2$ as an indicator of superfluous information to push down the corresponding element (i.e. ${\mathbf{h}_{j}^{l}}$) of hidden vectors to be zero with the confirmation that ${{\boldsymbol{\alpha}}_{j}^{l}=0}$ indicates the element does not carry any relevant information about the target ${\mathbf{y}}$ and can be pruned (which has been proved in \cite{dai2018compressing}).
Different from the methods that utilize sparsity-prompting regularization for CNN and LSTM layers \cite{dai2018compressing, Srivastava_2021_WACV}, we employ it to large pre-trained language models with Transformer-based architecture such as BERT for self-pruning purpose.

Finally, we take the first \textsc{[CLS]} token representation $\tilde{\mathbf{h}}^{L}_{\textsc{[CLS]}}$ of the compressed representations $\tilde{\mathbf{H}}^{L}$ from the BERT encoder and concatenate it with the output syntactic-aware representation $\tilde{\mathbf{h}}^{rel}_a$ of the target aspect from the
R-GAT network to form the final representation of the self-pruned network as $\tilde{\mathbf{h}}^{F}$.

\subsection{Self-pruning Contrastive Learning}
We design a self-pruning contrastive loss to optimize the two networks and improve the separability of all the classes, which narrows the distance between the representations of each anchor produced by the self-pruned and original networks while pushing apart the distance between the representations of different instances within a batch \cite{chen2020simple,khosla2020supervised,gao2021simcse}. In this way, the learned pruned network is expected to extract essential semantically relevant features and effectively generalize to long-tail instances by learning the inherent class characteristics. 

Specifically, for an anchor $x_{i}$, we first obtain its representations  ${\left(\mathbf{h}^{F}_{i},\tilde{\mathbf{h}}^{F}_{i}\right)}$ learned by the original and self-pruned network respectively as a positive pair. Meanwhile, the representations ${\left(\mathbf{h}^{F}_{i},\{\tilde{\mathbf{h}}^{F}_{j}\}_{j\neq i}^{N_m}\right)}$ of different input instances $\mathbf{x}_{i}$ and $\mathbf{x}_{j}$ within a mini-batch are treated as a negative pair.
Here, ${N_m}$ is the size of the mini-batch. The objective function for self-pruning contrastive learning can be formally defined as follows:
\begin{flalign}
     & \mathcal{L}_{SCL}\left(\theta_1,\theta_2\right) = -\frac{1}{N_m}\sum_{i=1}^{N_m}\,\mathrm{log}\left( \frac{sim\left(\mathbf{h}^{F}_{i},{\tilde{\mathbf{h}}^{F}_{i}};\tau\right)}{\sum_{j=1, i\neq j}^{N_m}{sim\left(\mathbf{h}^{F}_{i},{\tilde{\mathbf{h}}^{F}_{j}};\tau\right)}} \right) &  
\end{flalign}
where 
\begin{flalign}
& sim\left(\mathbf{h}^{F}_{i},{\tilde{\mathbf{h}}^{F}_{i}};\tau\right)=\mathrm{exp}\left({cosine}\left(\mathbf{h}^{F}_{i},{\tilde{\mathbf{h}}^{F}_{i}}\right)/\tau\right) &
\end{flalign}
where $sim\left(\cdot\right)$ is the similarity metric function. $cosine\left(\cdot\right)$ indicates the cosine similarity between two representations. ${\tau}$ is the temperature hyper-parameter.

\subsection{Joint Training Objective}
We jointly train the original network ${\mathcal{M}_{\theta_1}}$ and the self-pruned network ${\mathcal{M}_{\theta_2}^{(p)}}$ in an iterative manner until convergence. At each iteration, we train the original network with objective ${\mathcal{L}_{1}}$, which combines the cross-entropy loss with self-pruning contrastive loss. Similarly, we train the self-pruned network with the objective ${\mathcal{L}_{2}^{(p)}}$ which combines the VIB-based loss with the self-pruning contrastive loss. Mathematically, the objective functions ${\mathcal{L}_{1}}$ and ${\mathcal{L}_{2}^{(p)}}$ are defined as follows:
\begin{flalign}
    & \mathcal{L}_{1}=\mathcal{L}_{CE}\left(\theta_1\right)+{\gamma}\mathcal{L}_{SCL}\left(\theta_1,\theta_2\right) \\
    & \mathcal{L}_{2}^{(p)}=\mathcal{L}_{VIB}\left(\theta_2\right)+{\gamma}\mathcal{L}_{SCL}\left(\theta_1,\theta_2\right) &
\end{flalign}
\noindent where ${\mathcal{L}_{CE}}$, ${\mathcal{L}_{VIB}}$, ${\mathcal{L}_{SCL}}$ represent the cross-entropy loss, the VIB-based loss, and the self-pruning contrastive loss, respectively. $\gamma$ is a hyper-parameter to control the weights of the self-pruning contrastive loss for the original and self-pruned networks. Here, we set the value of $\gamma$ to ${0.25}$ in our experiments.

For the self-pruned network, at the feed-forward stage, we sample ${\epsilon^{l}}$ from $\mathcal{N}\left(0,I\right)$ for masks ${\mathbf{z}^l}$ and compute ${\tilde{\mathbf{H}}^{l}}$ across all BERT layers ${l}$. At the back-propagation stage, the overall parameters, including $\mu^{l}$ and $\sigma^{l}$ will be updated.

\begin{table}[t!]
\caption{\label{tab:datasets} Statistics of the ABSA datasets.}
    \centering
    \resizebox{1.\columnwidth}{!}{
        \begin{tabular}{l|c c c c}
        \specialrule{0.9pt}{0pt}{3.0pt}
        \bf Dataset & \bf Division & \bf \#\,Positive   & \bf \#\,Neutral & \bf \#\,Negative  \\
        \specialrule{0em}{0pt}{1.5pt}
        \hline
        \specialrule{0em}{0pt}{1.5pt}
        \multirow{2}{*}{\bf {REST4}} & Train & 2164 & 637 & 807  \\
        & Test &  728 & 196 & 196 \\
        \specialrule{0em}{0pt}{1.0pt}
        \hline
        \specialrule{0em}{0pt}{1.5pt}
        \multirow{2}{*}{\bf {LAP14}} & Train & 994 & 464 & 870  \\
        & Test & 341 & 169 & 128 \\
        \specialrule{0em}{0pt}{1.0pt}
        \hline
        \specialrule{0em}{0pt}{1.5pt}
        \multirow{2}{*}{\bf {REST15}} & Train & 912 & 36 & 256  \\
        & Test & 326 & 34 & 182 \\
        \specialrule{0em}{0pt}{1.5pt}
        \hline
        \specialrule{0em}{0pt}{1.pt}
        \multirow{2}{*}{\bf {REST16}} & Train & 1240 & 69 & 439  \\
        & Test & 469 & 30 & 117 \\
        \specialrule{0em}{0pt}{1.5pt}
        \hline
        \specialrule{0em}{0pt}{1.pt}
        \multirow{3}{*}{\bf {MAMS}} & Train & 3380 & 5042 & 2764  \\
        & Dev & 403 & 604 & 325 \\
        & Test & 400 & 607 & 329 \\
        \specialrule{0em}{0pt}{1.0pt}
        \hline
        \specialrule{0em}{0pt}{2.0pt}
        {{\bf REST14-ARTS}} & {Test} & {1953} & {473} & {1104} \\
        {{\bf LAP14-ARTS}} & {Test} & {883} & {407} & {587} \\
        \specialrule{0em}{0pt}{0.5pt}
        \specialrule{0.9pt}{0pt}{0pt}
        \end{tabular}}
\end{table}

\subsection{Inference Stage}
In the inference phase, given the back-propagation is disabled, we remove the original network and use the self-pruned network ${\mathcal{M}_{\theta_2}^{(p)}}$ to perform sentiment prediction directly. Note that at the $l$-th BERT layer, we merely use the mean vector ${\mu^{l}}$ without random sampling and mask the $j$-th element when the value of ${\boldsymbol{\alpha}^{l}_{j}}$ is zero.

\begin{table*}[t!]
\caption{\label{tab:main results} Main experimental results on five ABSA datasets. {The best scores are in bold, and the second-best ones are underlined. The results with ${\sharp}$ are retrieved from the corresponding original papers, and others are reported based on their released code and data.} 
}
\centering
\setlength{\tabcolsep}{7.0pt}
\renewcommand{\arraystretch}{1.2}
\resizebox{2.07\columnwidth}{!}{
\begin{tabular}{l cc cc cc cc cc}
    \hline
    \specialrule{0pt}{0pt}{0.3pt}
    \multirow{2}{*}{\textbf{Model}} & \multicolumn{2}{c}{\textbf{REST14 (${\%}$)}}   & \multicolumn{2}{c}{\textbf{LAP14 (${\%}$)}} & \multicolumn{2}{c}{\textbf{REST15 (${\%}$)}} & \multicolumn{2}{c}{\textbf{REST16 (${\%}$)}} & \multicolumn{2}{c}{\textbf{MAMS (${\%}$)}} \\ 
    \specialrule{0pt}{0.5pt}{0pt}
    \cline{2-11}
    \specialrule{0pt}{0pt}{0.5pt}
    & \textbf{Acc.} & \textbf{F1} & \textbf{Acc.} & \textbf{F1} & \textbf{Acc.} & \textbf{F1} & \textbf{Acc.} & \textbf{F1} & \textbf{Acc.} & \textbf{F1} \\
        \specialrule{0pt}{0pt}{0.3pt}
        \hline
        \specialrule{0pt}{1.0pt}{0pt}
            ATAE-LSTM \cite{wang2016attention} & 77.20$^{\sharp}$ & 67.02 & 68.70$^{\sharp}$ & 63.93 & 78.48 & 60.53 & 83.77 & 61.71 & {72.60}
            & {71.67} \\
            MemNet \cite{tang-etal-2016-aspect} &  80.95$^{\sharp}$ & 69.64 & 72.37$^{\sharp}$ & 65.17 & 77.31 & 58.28 & 85.44 & 65.99 & {67.89} 
            & {67.29} \\
            IAN \cite{ma2017interactive} & 78.60$^{\sharp}$ & 70.09 & 72.10$^{\sharp}$ & 67.38 & 78.54 & 52.65 & 84.74 & 55.21 & {73.13}
            & {72.53} \\
            MGAN \cite{fan2018multi} & 81.25$^{\sharp}$ & 71.94$^{\sharp}$ & 75.39$^{\sharp}$ & 72.47$^{\sharp}$ & 79.36 & 57.26 & 87.06 & 62.29 & {74.40} & {73.34} \\
            TNet \cite{li-etal-2018-transformation} & 80.69$^{\sharp}$ & 71.27$^{\sharp}$ & 76.54$^{\sharp}$ & 71.75$^{\sharp}$ & 78.47 & 59.47 & 89.07 & 70.43 & {75.15} & {74.45} \\
        \specialrule{0pt}{0pt}{0.3pt}
        \hline
        \specialrule{0pt}{1.0pt}{0pt}
           ASGCN \cite{zhang-etal-2019-aspect} & 80.86$^{\sharp}$ & 72.19$^{\sharp}$ & 74.14$^{\sharp}$ & 69.24$^{\sharp}$ & 79.34$^{\sharp}$ & 60.78$^{\sharp}$ & 88.69$^{\sharp}$ & 66.64$^{\sharp}$ & {75.82} & {74.19} \\
            CDT \cite{sun2019aspect} & 82.30$^{\sharp}$ & 74.02$^{\sharp}$ & 77.19$^{\sharp}$ & 72.99$^{\sharp}$ & - & - & 85.58$^{\sharp}$ & 69.93$^{\sharp}$ & {74.25} & {73.02} \\
            R-GAT \cite{wang2020relational} & 83.30$^{\sharp}$ & 76.08$^{\sharp}$ & 77.42$^{\sharp}$ & 73.76$^{\sharp}$ & 80.83 & 64.17 & 88.92 & 70.89 & {77.25} & {75.99} \\
            BiGCN \cite{zhang-qian-2020-convolution} & 81.97$^{\sharp}$ & 73.48$^{\sharp}$ & 74.59$^{\sharp}$ & 71.84$^{\sharp}$ & 81.16$^{\sharp}$ & 64.79$^{\sharp}$ & 88.96$^{\sharp}$ & 70.84$^{\sharp}$ & - & - \\
            Sentic GCN \cite{LIANG2022107643} & 84.03$^{\sharp}$ & 75.38$^{\sharp}$ & 77.90$^{\sharp}$ & 74.71$^{\sharp}$ & 82.84$^{\sharp}$ & 67.32$^{\sharp}$ & 90.88$^{\sharp}$ & 75.91$^{\sharp}$ & {76.57} & {75.56} \\
        \specialrule{0pt}{0pt}{0.3pt}
        \hline
        \specialrule{0pt}{1.0pt}{0pt}
            BERT-SPC \cite{song2019attentional} & 84.11 & 76.68 & 77.59 & 73.28 & 83.48 & 66.18 & 90.10 & 74.16 & {83.98} & {83.41} \\
            BERT-PT \cite{xu-etal-2019-bert} & 84.95$^{\sharp}$ & 76.96$^{\sharp}$ & 78.07$^{\sharp}$ & 75.08$^{\sharp}$ & - & - & - & - & - & - \\
            CapsNet-BERT \cite{jiang-etal-2019-challenge} & 85.36 & 78.41 & 78.97 & 75.66 & 82.10 & 65.57 & 90.10 & 75.15 & {83.76} & {83.15} \\
            TGCN-BERT \cite{tian-etal-2021-aspect} & 86.16$^{\sharp}$ & 79.95$^{\sharp}$ & \underline{80.88}$^{\sharp}$ & \underline{77.03}$^{\sharp}$ & \underline{85.26}$^{\sharp}$ & \underline{71.69}$^{\sharp}$ & \underline{92.32}$^{\sharp}$ & {77.29}$^{\sharp}$ & {83.38}$^{\sharp}$ & {82.77}$^{\sharp}$ \\
            \hdashline
            ASGCN\cite{zhang-etal-2019-aspect}-BERT & 85.54 & 78.54 & 77.90 & 73.92 & 82.47 & 68.83 & 91.07 & 76.13 & {83.46} & {83.00}
             \\
            \hspace{0.3cm}\textbf{w/ CVIB (Ours)} & {86.43} & {80.02} & {80.41} & {76.62} & {84.50} &	{71.42} & {91.72} & \underline{{78.95}} & \underline{{84.28}} & \underline{{83.87}} \\
            \hdashline
            RGAT-BERT \cite{wang2020relational} & \underline{86.60}$^{\sharp}$	& \underline{81.35}$^{\sharp}$ & 78.21$^{\sharp}$ & 74.07$^{\sharp}$ & 83.22 & 69.73 & 89.71 & 76.62 & 82.71 & 82.21 \\
            \hspace{0.3cm}\textbf{w/ CVIB (Ours)} & \textbf{87.59} & \textbf{82.03} & \textbf{81.35} & \textbf{77.53} & \textbf{85.98} & \textbf{72.84} & \textbf{92.86} & \textbf{82.87} & \textbf{84.66} & \textbf{84.03} \\
        \specialrule{0pt}{0pt}{0.5pt}
        \hline
\end{tabular}}
\end{table*}

\section{Experimental Setup}
\subsection{Datasets}
We conduct our experiments on five widely used benchmark datasets: \textbf{REST14} and \textbf{LAP14} from \cite{pontiki-etal-2014-semeval}, \textbf{REST15} from \cite{pontiki-etal-2015-semeval}, \textbf{REST16} from \cite{pontiki-etal-2016-semeval}, and \textbf{MAMS} from \cite{jiang-etal-2019-challenge}. We adopt the official data splits, which are the same as in the original papers. To test the robustness of ABSA models, we incorporate the \textbf{ARTS} datasets \cite{xing-etal-2020-tasty}, including REST14-ARTS and LAP14-ARTS, which extend the original REST14 and LAP14 testing datasets by applying three adversarial strategies: reversing the original sentiment of the target aspect (REVTGT), reversing the sentiment of the non-target aspects (REVNON), and generating more non-target aspects with opposite sentiment polarities from the target aspect (ADDDIFF). Each instance in these datasets consists of a review sentence, a target aspect, and the sentiment polarity (i.e., \textsc{Positive}, \textsc{Negative}, \textsc{Neutral}) towards the target aspect. The statistics of these used datasets are shown in Table \ref{tab:datasets}.

\subsection{Implementation Details}
In the experiments, we adopt the official pre-trained uncased BERT-base\footnote{https://github.com/huggingface/transformers}, which has ${12}$ layers and ${768}$ hidden dimensions. For the R-GAT network, we tune the number of relational self-attention heads varying from ${5}$ to ${8}$, and the other parameters follow the default configuration of the original paper \cite{wang2020relational}. For the VIB-based self-pruning loss, we set ${\beta^{l} =  1.0}$ for all layers, which is a simple yet effective choice in practice. The learnable vector $\mu^l$ is randomly initialized from a  distribution ${\mathcal{N}\left(1, 0.01\right)}$ and the logarithm of $\sigma^l$ is sampled from ${\mathcal{N}\left(-9, 0.01\right)}$. For the self-pruning contrastive loss, we set the hyper-parameter ${\tau=0.05}$. 
We train all our models for 30 epochs with Adam optimizer, and the initial learning rate is $0.00005$. The learning rate for the parameters of the VIB-based masking is initialized, varying from $0.001$ to $0.01$. 

\subsection{Baselines and Evaluation Metrics}
We compare the proposed CVIB method with three kinds of strong baselines, including \textbf{the attention-based methods}: ATAE-LSTM \cite{wang2016attention},  MemNet \cite{tang-etal-2016-aspect}, IAN \cite{ma2017interactive}, MGAN \cite{fan2018multi}, TNet \cite{li-etal-2018-transformation};  \textbf{the graph-based methods}: ASGCN \cite{zhang-etal-2019-aspect}, CDT \cite{sun2019aspect}, R-GAT \cite{wang2020relational}, BiGCN \cite{zhang-qian-2020-convolution}, Sentic GCN \cite{LIANG2022107643}; \textbf{the BERT-based methods}: BERT-SPC \cite{song2019attentional}, BERT-PT \cite{xu-etal-2019-bert}, CapsNet-BERT \cite{jiang-etal-2019-challenge}, RGAT-BERT \cite{wang2020relational}, TGCN-BERT \cite{tian-etal-2021-aspect}, ASGCN-BERT \cite{zhang-etal-2019-aspect}.

To evaluate the performance of the ABSA models, we adopt two widely used metrics: Accuracy (\textbf{Acc.}) and macro-averaged F1 score (\textbf{F1}). To ensure the stability of our CVIB method, we run CVIB ten times with random initialization and report the averaged results.

\begin{table*}[t!]
\caption{\label{tab:ablation study} Experimental results of ablation study on five datasets. ``VIB'' represents VIB-based pruning and ``SCL'' represents self-pruning contrastive learning.}
\centering
\setlength{\tabcolsep}{7.8pt}
\renewcommand{\arraystretch}{1.15}
\resizebox{1.8\columnwidth}{!}{
        \begin{tabular}{l cc cc cc cc cc}
        \hline
        \specialrule{0pt}{0pt}{0.3pt}
        \multirow{2}{*}{\bf Model} & \multicolumn{2}{c}{\textbf{REST14 (${\%}$)}}   & \multicolumn{2}{c}{\textbf{LAP14 (${\%}$)}} & \multicolumn{2}{c}{\textbf{REST15 (${\%}$)}} & \multicolumn{2}{c}{\textbf{REST16 (${\%}$)}} & \multicolumn{2}{c}{\textbf{MAMS (${\%}$)}} \\ 
        \specialrule{0pt}{0.5pt}{0pt}
        \cline{2-11} 
        \specialrule{0pt}{0pt}{0.5pt}
        & \textbf{Acc.} & \textbf{F1} & \textbf{Acc.} & \textbf{F1} & \textbf{Acc.} & \textbf{F1} & \textbf{Acc.} & \textbf{F1} & \textbf{Acc.} & \textbf{F1} \\
        \specialrule{0pt}{0.5pt}{0pt}
        \hline
        \specialrule{0pt}{0pt}{0.5pt}
            \textbf{CVIB} & \textbf{87.59} & \textbf{82.03} & \textbf{81.35} & \textbf{77.53} & \textbf{85.98} & \textbf{72.84} & \textbf{92.86} & \textbf{82.87} & \textbf{84.66} & \textbf{84.03} \\
            \hspace{0.3cm}{w/o} VIB & 84.82 & 78.70 & 78.97 & 74.44 & 84.50 & 67.05 & 89.45 & 76.14  & 83.28 & 83.81 \\
            \hspace{0.3cm}{w/o} SCL & 85.63 & 78.32 & 79.94 & 76.62 & 85.24 & 71.80 & 91.56	 & 77.74 & 84.36 & 83.93 \\
            \hspace{0.3cm}{{w/o}} {VIB+SCL} & {84.38} & {77.71} & {78.37} & {73.45} & {83.22} & {69.73} & {89.71} & {76.62} & {82.71} & {82.21} \\
        \specialrule{0pt}{0pt}{0.5pt}
        \hline
        \end{tabular}}
\end{table*}

\section{Experimental Results}
\subsection{Main Results}
The experimental results of the ABSA methods on five benchmark datasets are reported in Table \ref{tab:main results}. We observe that CVIB achieves the best performance against all baselines on the five datasets. 
The performance of attention-based ABSA methods (e.g., IAN, MGAN, TNet) is comparatively lower than that of graph-based and BERT-based methods. This discrepancy arises from the inherent limitations of attention mechanisms, which implicitly model relationships between the target aspect and its context. These mechanisms may learn incorrect associations or attend to irrelevant features for the target aspect, thus limiting the performance of ABSA. 
In contrast, graph-based methods leverage syntactic dependencies to explicitly model the aspect-context relationships, while BERT-based methods utilize pre-trained language models (PLMs) to learn contextually distinguishable representations for the target aspect. 
Notably, TGCN-BERT and RGAT-BERT outperform the other baselines by taking benefits of both the graph knowledge based on the syntactic dependencies and the rich linguistic knowledge contained in PLMs.
Our proposed CVIB performs even better than the best-performing baseline RGAT-BERT, which is the backbone of our original network, in terms of all evaluation metrics, verifying the effectiveness of the CVIB framework. 
Furthermore, CVIB achieves consistent and substantial improvements when integrated with another strong baseline, namely ASGCN-BERT. These advancements demonstrate that our proposed CVIB framework can reduce spurious correlations between input features and output labels, thus enhancing the performance of the ABSA models. 

\subsection{Ablation Study}
To investigate the impact of different components on the overall performance of our proposed method, we conduct an ablation study on the five ABSA datasets. Concretely, we perform two ablations: (1) removing the VIB-based pruning (denoted as ``w/o VIB'') from CVIB by employing a random dropout strategy to train the self-pruned network; (2) removing the self-pruning contrastive learning (denoted as ``w/o SCL'') by training a single network based on the VIB principle.

The ablation test results are reported in Table \ref{tab:ablation study}. The performance of CVIB drops sharply when discarding the VIB-based pruning. 
This aligns with our expectation since the VIB-based pruning enables the classifier to reduce spurious correlations between input features and output prediction, thus improving the robustness and generalization capability of the classifier. 
SCL also makes a considerable contribution to CVIB despite the slightly inferior results on certain datasets (e.g., REST16). One potential explanation is that solely employing SCL may lead to the reliance on spurious correlations for distinguishing different sentiment classes.
It is no surprise that combining VIB-based pruning and SCL contributes to a significant improvement in CVIB. VIB can reduce spurious correlations between input features and output labels, and then SCL can further capture semantically relevant features from the learned representations, thus improving the separability of all the classes.

\begin{table}[t!]
\caption{\label{tab:robustness results} Robustness results on aspect robustness test sets ({ARTS}). We compare accuracy ({Acc.}) and macro-averaged F1 ({F1}) on the original and the new test sets. We also calculate the performance drops ({Drop}) from original sets to perturbed sets.}
\begin{center}
\setlength{\tabcolsep}{3.3pt}
\renewcommand{\arraystretch}{1.3}
\resizebox{1.0\columnwidth}{!}{
        \begin{tabular}{l cc cc}
        \hline
        \specialrule{0pt}{0pt}{0.3pt}
        \textbf{Model} & \textbf{Acc. 
 (Ori${\to}$New) } & \textbf{Drop
        } &  \textbf{F1 (Ori${\to}$New)} & \textbf{Drop}  \\
        \specialrule{0pt}{0.5pt}{0pt}
        \hline
        \specialrule{0pt}{0pt}{0.5pt}
        \multicolumn{5}{c}{\textbf{REST14-ARTS (${\%}$)}} \\
        \specialrule{0pt}{0.5pt}{0pt}
        \hline
        \specialrule{0pt}{0pt}{0.5pt}
        ATAE-LSTM & 77.20 ${\to}$ 58.45 & -18.75 & 67.02 ${\to}$ 49.65 & -17.37 \\ 
        MemNet & 79.61 ${\to}$ 55.30 & -24.31 & 69.64 ${\to}$ 46.67 & -22.97 \\
        IAN & 79.26 ${\to}$ 57.75 & -21.51 & 70.09 ${\to}$ 48.12 & -21.97 \\
        ASGCN & 80.86 ${\to}$ 59.12 & -21.74 & 72.19 ${\to}$ 35.56 & -36.63 \\
        R-GAT & 83.30 ${\to}$ 60.31 & -22.99 & 76.08 ${\to}$ 37.51 & -38.57 \\
        BERT-SPC & 84.11 ${\to}$ 57.81 & -26.30 & 76.68 ${\to}$ 48.08 & -28.60 \\
        CapsNet-BERT & 85.36 ${\to}$ 69.24 & -16.12 & 78.41 ${\to}$ 55.25 & -23.16 \\
        RGAT-BERT & 86.60 ${\to}$ 71.64 & -14.96 & 81.35 ${\to}$ 60.10 & -21.25 \\
        \textbf{CVIB (Ours)} & \textbf{87.59} ${\to}$ \textbf{77.48} & \textbf{-10.11} & \textbf{82.03} ${\to}$ \textbf{71.74} & \textbf{-10.29} \\
        \specialrule{0pt}{0.5pt}{0pt}
        \hline
        \specialrule{0pt}{0pt}{0.5pt}
        \multicolumn{5}{c}{\textbf{LAP14-ARTS (${\%}$)}}\\
        \specialrule{0pt}{0.5pt}{0pt}
        \hline
        \specialrule{0pt}{0pt}{0.5pt}
        ATAE-LSTM & 68.70 ${\to}$ 51.33 & -17.37 &  63.93 ${\to}$ 46.11 & -17.82 \\
        MemNet & 70.64 ${\to}$ 52.00 & -18.64 & 65.17 ${\to}$ 46.50 & -18.67 \\
        IAN & 72.05 ${\to}$ 52.91 & -19.14 & 67.38 ${\to}$ 47.54 & -19.84 \\
        ASGCN & 75.55 ${\to}$ 49.81 & -25.74 & 71.05 ${\to}$ 36.45 & -34.60 \\
        R-GAT & 77.42 ${\to}$ 50.61 & -26.81 & 73.76 ${\to}$ 36.10 & -37.66 \\
        BERT-SPC & 77.59 ${\to}$ 58.02 & -19.57 & 73.28 ${\to}$ 54.58 & -18.70 \\
        CapsNet-BERT & 78.97 ${\to}$ 60.31 & -18.66 & 75.66 ${\to}$ 51.50 & -24.16 \\
        RGAT-BERT & 78.21 ${\to}$ 66.30 & -11.91 & 74.07 ${\to}$ 55.68 & -18.39 \\
        \textbf{CVIB (Ours)} & \textbf{81.35} ${\to}$ \textbf{73.21} & \textbf{\hspace{0.2cm}-8.14} & \textbf{77.53} ${\to}$ \textbf{70.47} & 
        \textbf{\hspace{0.2cm}-7.06} \\
        \specialrule{0pt}{0pt}{0.5pt}
        \hline
        \end{tabular}}
\end{center}
\end{table}

\begin{table}[t!]
\caption{ \label{tab:arts subset results} {The robustness results on ARTS subsets of three different adversarial strategies (i.e., REVTGT, REVNON, ADDDIFF).}}
\centering
\setlength{\tabcolsep}{13.8pt}
\renewcommand{\arraystretch}{1.2}
\resizebox{1.0\columnwidth}{!}{
        \begin{tabular}{l cc cc}
        \hline
        \specialrule{0pt}{0pt}{0.3pt}
        \multirow{2}{*}{{\bf Set}} & \multicolumn{2}{c}{{\bf REST14-ARTS (${\%}$)}} & \multicolumn{2}{c}{{\bf LAP14-ARTS (${\%}$)}}  \\ 
        \specialrule{0pt}{0.5pt}{0pt}
        \cline{2-5} 
        \specialrule{0pt}{0pt}{0.5pt}
        & {\bf Acc.} & {\bf F1} & {\bf Acc.} & {\bf F1} \\
        \specialrule{0pt}{0.5pt}{0pt}
        \hline
        \specialrule{0pt}{0pt}{0.5pt}
        {FULL} & {77.48} & {71.74} & {73.21} & {70.47} \\
        {REVTGT} & {68.32} & {56.16} & {60.09} & {52.18} \\
        {REVNON} & {74.32} & {57.54} & {78.52} & {62.87} \\
        {ADDDIFF} & {79.20} & {71.79} & {74.92} & {69.99} \\
        \specialrule{0pt}{0pt}{0.5pt}
        \hline
        \end{tabular}}
\end{table}
\subsection{Robustness Analysis}
We evaluate the robustness of CVIB on \textbf{Aspect Robustness Test Sets (ARTS)} \cite{xing-etal-2020-tasty}, which are constructed to test whether a model can robustly capture the aspect-relevant information to distinguish the sentiment towards the target aspect from the non-target aspects. ARTS extends the original test sets of REST14 and LAP14 corpora by applying three adversarial strategies: reversing the original sentiment of the target aspect (REVTGT), reversing the sentiment of the non-target aspects (REVNON), and generating more non-target aspects with opposite sentiment polarities from the target aspect (ADDDIFF). Since CVIB focuses on reducing spurious correlations (e.g., irrelevant information) from the non-target aspects and captures truly target-relevant sentiment information, we assume that CVIB will show strong robustness in adversarial scenarios.

The results are shown in Table \ref{tab:robustness results}. We observe that CVIB achieves substantially better performance than the compared methods when injecting adversarial perturbations, which verifies the robustness of our proposed CVIB framework. For example, on the REST14-ARTS dataset, the overall accuracy and F1 scores drop 10.31\% and 10.50\%, which are much better than that produced by RGAT-BERT (i.e., 14.96\% and 21.25\%).

In addition, we evaluate the robustness of our proposed method on three perturbed subsets (i.e., REVTGT, REVNON and ADDDIFF), respectively. As shown in Table \ref{tab:arts subset results}, CVIB demonstrates commendable efficacy on the REVNON and ADDDIFF subsets. This underscores the inherent capacity of CVIB to mitigate spurious correlations and effectively capture semantically relevant features pertaining to the target aspect. Such proficiency contributes to heightened robustness against inconsequential features associated with non-target aspects. However, CVIB exhibits comparatively diminished performance on the REVTGT subset. This particular challenge emanates from the inherent complexity faced by ABSA models in accurately discerning the reversed sentiment polarity of the target aspect amidst subtle textual modifications.

\begin{table}[t!]
\caption{ \label{tab:OOD results} The cross-domain results for verifying the generalization capability of the ABSA models. Here, ``{REST14${\to}$LAP14}'' indicates that  training the model on {REST14} and testing it on {LAP14}.}
\centering
\setlength{\tabcolsep}{7.5pt}
\renewcommand{\arraystretch}{1.2}
\resizebox{1.0\columnwidth}{!}{
        \begin{tabular}{l cc cc}
        \hline
        \specialrule{0pt}{0pt}{0.3pt}
        \multirow{2}{*}{\textbf{Model}} & \multicolumn{2}{c}{\textbf{REST14${\to}$LAP14 (${\%}$)}}   & \multicolumn{2}{c}{\textbf{LAP14${\to}$REST14 (${\%}$)}}  \\ 
        \specialrule{0pt}{0.5pt}{0pt}
        \cline{2-5} 
        \specialrule{0pt}{0pt}{0.5pt}
        & {\textbf{Acc.}} & {\textbf{F1}} &  {\textbf{Acc.}} & {\textbf{F1}} \\
        \specialrule{0pt}{0.5pt}{0pt}
        \hline
        \specialrule{0pt}{0pt}{0.5pt}
        ATAE-LSTM  & 62.85 & 54.97 & 71.61 & 53.61 \\
        MemNet  & 57.84	& 51.15 & 70.66	& 52.07 \\
        IAN  & 63.82 & 55.20 & 72.09 & 54.44 \\
        ASGCN  & 60.34	& 42.54 & 71.52	& 46.12 \\
        R-GAT & 61.76	& 44.17 & 70.18	& 45.81 \\
        BERT-SPC & 70.06 & 61.53 & 75.89 & 65.47\\
        CapsNet-BERT & 68.97 & 59.18 & 77.41 & 56.55 \\
        RGAT-BERT & 74.27	& 70.38 & 79.07	& 70.32 \\
        \textbf{CVIB (Ours)} & \textbf{77.43} & \textbf{73.57} & \textbf{81.07} & \textbf{72.76} \\
        \specialrule{0pt}{0pt}{0.5pt}
        \hline
        \end{tabular}}
\end{table}

\begin{table}[t!]
\caption{\label{tab:long-tail} The performance of the ABSA models on two long-tail distributions. Here, {Positive} ({Pos.}) is the largest class and {Neutral} ({Neu.}) is the smallest class. }
\centering
\setlength{\tabcolsep}{5.pt}
\renewcommand{\arraystretch}{1.25}
    \resizebox{1.0\columnwidth}{!}{
        \begin{tabular}{l ccc ccc}
        \specialrule{0.9pt}{0pt}{0pt}
        \multirow{2}{*}{\textbf{Model}} & \multicolumn{3}{c}{\textbf{REST15 (${\%}$)}} & \multicolumn{3}{c}{\textbf{REST16 (${\%}$)}} \\ \specialrule{0pt}{0.5pt}{0pt}
        \cline{2-7} 
        \specialrule{0pt}{0pt}{0.5pt}
        & \textbf{Pos.} & \textbf{Neg.} & \textbf{Neu.} & \textbf{Pos.} & \textbf{Neg.} & \textbf{Neu.} \\
        \specialrule{0pt}{0.5pt}{0pt}
        \hline
        \specialrule{0pt}{0pt}{0.5pt}
        BERT-SPC & 92.10  & 81.52 & 11.76 & 96.10 & 80.25 & 33.33 \\
        CapsNet-BERT & 91.12  & 79.36 & 11.76 & 95.74 & 83.76 & 30.00  \\
        RGAT-BERT & 91.03  & 81.50 & 17.65 & 95.95 & 84.62 & 13.33 \\
        \textbf{CVIB (Ours)} & \textbf{92.80} & \textbf{85.16} & \textbf{26.47} & \textbf{96.59} & \textbf{88.03} & \textbf{53.33}  \\
        \specialrule{0pt}{0pt}{0.5pt}
        \hline
        \end{tabular}}
\end{table}

\subsection{Generalization Analysis}
We first evaluate the generalization capability of our proposed CVIB in the cross-domain scenario.
Concretely, we train our CVIB on the REST14 training set (in the restaurant domain) and then test CVIB on the LAP14 testing set (in the laptop domain). Similarly, we also train CVIB on the LAP14 training set and then test CVIB on the REST14 testing set. 
The experimental results are reported in Table \ref{tab:OOD results}. We observe that CVIB outperforms the compared methods by a large margin, which verifies the outstanding generalization capability of our proposed CVIB framework. CVIB is expected to learn more transferable features and thus achieve better generalization capability than the compared methods in the cross-domain scenario.

\subsection{Performance on Long-tail Samples}
We also evaluate the generalization performance of CVIB in the long-tail scenario. As shown in Table \ref{tab:datasets}, for both REST15 and REST16 datasets, the class size of the Positive class (the largest class) divided by the Neutral class (the smallest class) is more than 10. 
In Table \ref{tab:long-tail}, we report the averaged prediction results of the three classes (i.e., Positive, Negative, and Neutral) separately. Our CVIB method achieves a substantially better performance of the minority class (i.e., Neutral) than the compared baselines. This verifies that CVIB can learn better representations for difficult-to-memorize samples and generalize well in the long-tail scenario. 

\begin{figure*}[t!]
\centering
\includegraphics[width=140mm]{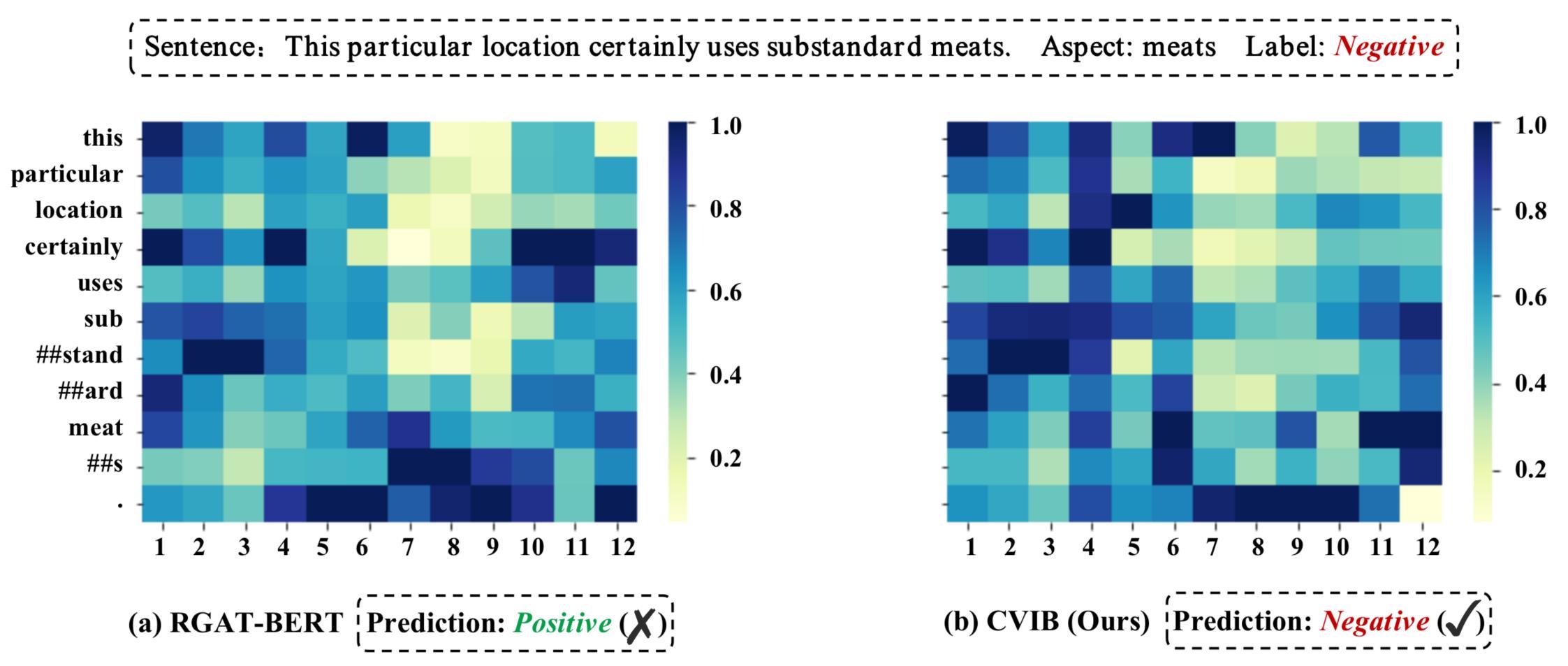}
\caption{\label{fig:case study} Visualization results for RGAT-BERT (a) and CVIB (b). Here, the x-axis represents the number of BERT layers from the 1st to 12th layers, and the y-axis represents the tokens of the example sentence.}
\end{figure*}

\subsection{Case Study}
We use a representative exemplary case that is selected from the \textsc{Rest14} testing set to further investigate the effectiveness of CVIB. This instance is incorrectly predicted by RGAT-BERT while being correctly predicted by CVIB. We visualize the self-attention scores across the BERT layers. 
As shown in Fig. \ref{fig:case study} (a), RGAT-BERT has a tendency to pay more attention to irrelevant words such as  ``particular'' and ``certainly'', which frequently co-occur with the \textsc{Positive} label in the training data.
RGAT-BERT suffers from statistically spurious correlations between input features and output prediction, failing to make a correct prediction. In Fig. \ref{fig:case study} (b), 
CVIB obtains the correct sentiment label by weakening the influence of task-irrelevant words and capturing useful semantically relevant words (e.g., ``substandard'') that carry important sentiment clues for label prediction. 

\section{Conclusion}
In this paper, we proposed a contrastive variational information bottleneck framework (called CVIB) to mitigate the spurious correlation problem for the ABSA task, improving the robustness and generalization capability of the deep ABSA method. CVIB is composed of an original network and a self-pruned network, which are learned simultaneously via contrastive learning. 
First, the self-pruned network was learned adaptively from the original network based on the VIB principle, which discarded the spurious correlations while preserving sufficient information about the sentiment labels.
Then, a self-pruning contrastive loss was devised to optimize the two networks and improve the separability of all the classes. Consequently, the self-pruned network reduced the spurious correlations, making it easier for the ABSA classifier to avoid overfitting. We conducted extensive experiments on five benchmark datasets, and the experimental results showed the effectiveness of CVIB.

\section*{Acknowledgments}
Min Yang was supported by the National Key Research and Development Program of China (2022YFF0902100), National Natural Science Foundation of China (62376262), Shenzhen Science and Technology Innovation Program (KQTD20190929172835662), Shenzhen Basic Research Foundation (JCYJ20210324115614039 and JCYJ20200109113441941). Qingshan Jiang was supported by National Key Research and Development Program of China (2021YFF1200100 and 2021YFF1200104).


\bibliographystyle{model1-num-names}
\bibliography{cvib}


\end{sloppypar}
\end{document}